\pdfoutput=1

\documentclass[11pt,a4paper]{article}

\usepackage[hyperref]{acl2021}
\aclfinalcopy

\usepackage{times}
\usepackage{latexsym}
\usepackage{lineno}

\usepackage[T1]{fontenc}

\usepackage[utf8]{inputenc}

\usepackage{microtype}

\usepackage{inconsolata}

\usepackage{graphicx}

\usepackage[whole]{bxcjkjatype}
\usepackage{lingmacros}
\usepackage{amsmath,amssymb}
\usepackage{mathtools}
\usepackage{ascmac}
\usepackage{proof,lscape,caption}
\usepackage{multirow}
\usepackage{color}
\usepackage{stmaryrd}
\usepackage{url}
\usepackage{booktabs}
\usepackage{natbib}
\bibpunct[:]{(}{)}{,}{a}{}{,}
\usepackage{pifont}
\usepackage{enumerate}
\usepackage[margin=5pt]{subcaption}
\usepackage{makecell}
\usepackage{bussproofs}
\usepackage{lscape}
\usepackage{bm}
\usepackage{longtable}
\usepackage{leipzig}
\usepackage{gb4e}
\noautomath
\usepackage{breqn}
\usepackage{cleveref}

    {\gdef\scalefactor{#1}\begin{center}\proofSkipAmount \leavevmode}%
    {\scalebox{\scalefactor}{\DisplayProof}\proofSkipAmount \end{center}%
}

\makeglossaries

\newleipzig{npst}{npst}{nonpast}
\newleipzig{qp}{qp}{questionparticle}
\newleipzig{no}{no}{nominalizer}

\newcommand{\systemname}{ccg-jcomp}

\title{Can Large Language Models Robustly Perform \\
Natural Language Inference for Japanese Comparatives?}

\author{
    Yosuke Mikami${}^{1, 2}$ \quad
    Daiki Matsuoka${}^{1, 2}$ \quad
    Hitomi Yanaka${}^{1, 2}$ \\
    ${}^1$The University of Tokyo\\
    ${}^2$Riken \\
    \texttt{\{ymikami, daiki.matsuoka, hyanaka\}@is.s.u-tokyo.ac.jp}
}

\begin{document}
\maketitle
\begin{abstract}
Large Language Models (LLMs) perform remarkably well in Natural Language Inference (NLI).
However, NLI involving numerical and logical expressions remains challenging.
Comparatives are a key linguistic phenomenon related to such inference, but the robustness of LLMs in handling them, especially in languages that are not dominant in the models’ training data, such as Japanese, has not been sufficiently explored.
To address this gap, we construct a Japanese NLI dataset that focuses on comparatives and evaluate various LLMs in zero-shot and few-shot settings.
Our results show that the performance of the models is sensitive to the prompt formats in the zero-shot setting and influenced by the gold labels in the few-shot examples.
The LLMs also struggle to handle linguistic phenomena unique to Japanese.
Furthermore, we observe that prompts containing logical semantic representations help the models predict the correct labels for inference problems that they struggle to solve even with few-shot examples.
\end{abstract}

\section{Introduction}
\label{chap:intro}
In recent years, Large Language Models (LLMs) have demonstrated high performance across a wide range of tasks, including Natural Language Inference (NLI;~\citealp{bowman-etal-2015-large}).
However, inference with numerical and logical expressions remains challenging for LLMs~\cite{she2023scone,liu2023evaluatinglogicalreasoningability,parmar-etal-2024-logicbench}.
In particular, NLI involving comparatives is important, as it requires a proper understanding of such expressions.
Indeed, there are English benchmarks focusing on comparatives for pre-trained models and inference systems~\citep{Haruta_Mineshima_Bekki_2022, liu2023adjective}.

However, it has not been thoroughly investigated how \textit{robust} LLMs are in handling various types of inference involving comparatives, regardless of the prompt formats or the few-shot example selection.
Moreover, 
there is growing attention to analyzing the robustness of inference in languages that are not dominant in the pre-training data.

Given these motivations, we construct an NLI dataset focusing on Japanese comparatives by creating templates from an existing Japanese NLI dataset and filling in them with words.\footnote{Our dataset is available on \url{https://github.com/ynklab/comparativeNLI_dataset}}
Using this dataset, we evaluate five LLMs, including both open and commercial models.
We analyze how robustly LLMs can perform inference on comparatives regardless of the way prompts are given in zero-shot and few-shot settings.
We also compare LLMs with \systemname{}\footnote{\url{https://github.com/ynklab/ccg-jcomp}}~\cite{ccgjcomp}, a logical inference system for Japanese comparatives.

The experimental results suggest that the prompt formats impact the model behavior in the zero-shot settings, and that the few-shot performance is influenced by the gold labels in the few-shot examples.
In addition, prompts with semantic representations from \systemname{} can improve model accuracy on problems that remain difficult even with standard few-shot settings.

\definecolor{mygreen}{RGB}{44, 160, 44}
\newcommand{\red}[1]{\textcolor{red}{#1}}
\newcommand{\blue}[1]{\textcolor{blue}{#1}}
\newcommand{\green}[1]{\textcolor{mygreen}{#1}}
\begin{table*}[t]
    \centering
    \small{
    \begin{tabular}{l|l|cl|l|l}
    \hline
        ID & Category & \multicolumn{2}{l|}{Template} & Example & Label \\
    \hline\hline
        \multirow{6}{*}{jsem-570} & %
        & \multirow{3}{*}{P} & \blue{X}-wa \green{Y}-yori \red{A}. & \blue{Taro}-wa \green{Hanako}-yori \red{omoi}. & \multirow{6}{*}{\textit{unk}} \\
         & & & \blue{X}-\Top{} \green{Y}-than \red{A} & \blue{Taro}-\Top{} \green{Hanako}-than \red{heavy} \\
         & basic & & (\blue{X} is more \red{A} than \green{Y}) & (\blue{Taro} is \red{heavier} than \green{Hanako}) \\
         \cline{3-5}
         & comparative & \multirow{3}{*}{H} & \blue{X}-wa \red{A}. & \blue{Taro}-wa \red{omoi}. & \\
         & & & \blue{X}-\Top{} \red{A} & \blue{Taro}-\Top{} \red{heavy} & \\
         & & & (\blue{X} is \red{A}) & (\blue{Taro} is \red{heavy}) &\\
    \hline
        \multirow{6}{*}{jsem-577} & \multirow{6}{*}{equative} & \multirow{3}{*}{P} & \blue{X}-wa \green{Y}-to onaji-kurai-no \red{$\text{N}_{\text{A}}$}-da. & \blue{Taro}-wa \green{Jiro}-to onaji-kurai-no \red{omosa}-da. & \multirow{6}{*}{\textit{unk}}\\
         & & & \blue{X}-\Top{} \green{Y}-\Com{} as \red{$\text{N}_{\text{A}}$}-\Cop{} & \blue{Taro}-\Top{} \green{Jiro}-\Com{} as \red{weight}-\Cop{} & \\
         & & & (\blue{X} is as \red{A} as \green{Y}) & (\blue{Taro} is as \red{heavy} as \green{Jiro}) &\\
         \cline{3-5}
         & & \multirow{3}{*}{H} & \blue{X}-wa \green{Y}-yori \red{A}. & \blue{Taro}-wa \green{Jiro}-yori \red{omoi}. & \\
         & & & \blue{X}-\Top{} \green{Y}-than \red{A} & \blue{Taro}-\Top{} \green{Jiro}-than \red{heavy} &\\
         & & & (\blue{X} is more \red{A} than \green{Y}) & (\blue{Taro} is \red{heavier} than \green{Jiro}) &\\
    \hline
        \multirow{6}{*}{jsem-620} & \multirow{6}{*}{presupposition} & \multirow{3}{*}{P} & \blue{X}-wa \green{Y} izyoo-ni \red{A}. & \blue{Taro}-wa \green{Hanako} izyoo-ni \red{omoi}. & \multirow{6}{*}{\textit{yes}}\\
         & & & \blue{X}-\Top{} \green{Y} than \red{A} & \blue{Taro}-\Top{} \green{Hanako} than \red{heavy} & \\
         & & & (\blue{X} is more \red{A} than \green{Y}) & (\blue{Taro} is \red{heavier} than \green{Hanako}) & \\
         \cline{3-5}
         & & \multirow{3}{*}{H} & \green{Y}-wa \red{A}. & \green{Hanako}-wa \red{omoi}. & \\
         & & & \green{Y}-\Top{} \red{A} & \green{Hanako}-\Top{} \red{heavy} & \\
         & & & (\green{Y} is \red{A}) & (\green{Hanako} is \red{heavy}) &\\
    \hline
    \end{tabular}
    }
    \caption{Examples of categories and their corresponding templates. P and H denote the premise and the hypothesis, respectively. \blue{X} (\green{Y}), \red{A}, and \red{$\text{N}_{\text{A}}$} are a proper noun, an adjective, and the noun form of an adjective, respectively.
    ID indicates the ID in the original JSeM dataset.
    \textit{unk} stands for the \textit{unknown} label.}
    \label{tab:dataset_example}
\end{table*}

\section{Related Work}
\label{chap:related}
In this section, we describe existing datasets that contain inference problems involving comparatives.
JSeM~\citep{10.1007/978-3-319-50953-2_5} is a Japanese NLI dataset, constructed from the English NLI dataset FraCaS~\citep{cooper1996using} with some additional problems that cover inference unique to Japanese.
The problems are divided into sections based on semantic phenomena, including comparatives, which allows us to evaluate the strengths and weaknesses of models with respect to individual phenomena.
However, since JSeM is limited in vocabulary and small in scale, we create templates from the dataset and generate new problems by filling in the templates with various words.

CAD~\citep{Haruta_Mineshima_Bekki_2022} is a dataset on English adjectives, comparatives, adverbs, and quantifiers.
The authors chose inference examples from linguistic papers and constructed new problems by applying transformations such as adding negation and replacing words.
Adjective Scale Probe~\citep{liu2023adjective} is a dataset designed to investigate how well language models understand degree semantics.
It is semi-automatically generated based on templates.
While these studies evaluate the extent to which pre-trained language models perform inference involving comparatives in fine-tuned settings, they do not specifically focus on the robustness of the inference in in-context learning settings.
To address this gap, we provide a scalable NLI dataset involving Japanese comparatives based on templates created from existing hand-crafted NLI problems.

\section{Dataset Creation}
\label{chap:dataset}
To analyze the extent to which LLMs robustly perform inference involving Japanese comparatives, we create an NLI dataset based on the comparatives section of JSeM.
Our dataset construction process is composed of (i) template creation based on JSeM and (ii) problem creation using the templates.

\subsection{Template Creation}
\label{subsec:category}

First, for each problem in JSeM, we manually construct a template containing blanks for adjectives, verbs, numerals, and nouns.
Each template has at least one premise and one hypothesis.
The gold labels are \textit{yes}, \textit{no}, and \textit{unknown}, corresponding to entailment, contradiction, and neutral, respectively.

The templates are classified into ten categories based on JSeM:
basic comparative, equative, clausal comparative, numerical, ambiguous, temporal, quantifier, absolute adjective, presupposition, and superlative.
One problem may have multiple categories.

\Cref{tab:dataset_example} shows some examples of categories and their corresponding templates.
In what follows, we will refer to a template with its original ID in JSeM, which is shown in the leftmost column.
First, jsem-570 involves a basic comparative expression \textit{yori}.
Second, jsem-577 targets the equative construction, with its premise meaning that the degree of property A is almost the same for X and Y.
Since the premise does not specify which degree is greater, its gold label is \textit{unknown}.
Third, jsem-620 is one of the problems focusing on the fact that some Japanese comparative expressions trigger a presupposition~\citep{kubota2012presuppositional,hayashishita2007izyoo}.
Here, the phrase ``izyoo-ni'' makes the premise presuppose that Y is A, as a result of which the premise entails the hypothesis.

\subsection{Problem Creation}
\label{sec:problem}
We create new problems by filling in the templates with words corresponding to each part of speech, in order to see whether the models can consistently capture the inference patterns independently of specific content words.
The words to be inserted into the templates are carefully chosen by the authors, who are native speakers of Japanese, for their naturalness.
In what follows, we detail the concrete procedure for word insertion.

As for a placeholder for an adjective, we insert gradable adjectives in a way that the gold label remains unchanged.
More specifically, we avoid using a certain class of adjectives called \textit{absolute adjectives} \citep{Kennedy2005ScaleSD}, which allow inference from ``X is more A than Y'' to ``Y is A'' (e.g., ``wet'').
Since this property may lead to undesirable changes to the gold label in some templates, we make sure that the inserted word is not an absolute adjective.

In addition, we adopt different strategies depending on whether the placeholder involves the predicative or attributive use.
With the predicative use, we insert only adjectives that can take a person as their subject.
When the placeholder for an adjective involves the attributive use, in which case the whole template also contains placeholders for a noun and a verb, we construct and apply a list of plausible adjective-noun-verb combinations.
More concretely, we first input the template into GPT-4o to generate some adjective-noun-verb combinations.
Then, we manually select natural ones from them.
To illustrate, consider the template ``Taro [verb] a more [adjective] [noun] than Jiro'' (for expository purposes, we write the template in English).
If the LLM produces the combinations \textit{expensive}-\textit{car}-\textit{bought} and \textit{expensive}-\textit{backpack}-\textit{drank}, we choose the first output but not the second, since only the first combination results in a semantically-natural sentence when inserted.

Finally, for templates involving numerals, we set a natural range of numerical values compatible with the lexical item for each problem and select the numbers to fill in the templates within that range.
For instance, in the template ``Taro ate [number] apples,'' we choose numbers less than 5.

With these strategies, we generate approximately 60 problems from each template.
As a result, the total number of problems is 4304, and the distribution of the gold labels is (\textit{yes}/\textit{no}/\textit{unknown}) = (2524/466/1314).

\section{Evaluation of Zero-shot NLI}
\label{chap:zeroshot}
First, we analyze how consistent the performance of the LLMs is regardless of the prompts in the zero-shot prompt setting, compared with a logical inference system.

\subsection{Experimental Setting}
\label{sec:settings1}

\paragraph{Models}
We evaluate five LLMs: GPT-4o\footnote{\url{https://openai.com/index/gpt-4o-system-card/}}, Llama-3.1-8B/70B\footnote{\url{https://huggingface.co/collections/meta-llama/llama-31-669fc079a0c406a149a5738f}} (Llama8B/70B), instruction-tuned Llama-3.1-8B/70B~\citep{grattafiori2024llama}, Llama-3.1-Swallow-8B/70B\footnote{\url{https://huggingface.co/collections/tokyotech-llm/llama-31-swallow-66fd4f7da32705cadd1d5bc6}} (Swallow8B/70B), and instruction-tuned Llama-3.1-Swallow-8B/70B~\citep{fujii2024continual}.
Llama 8B/70B are open-source and multilingual models but do not officially support Japanese.
Swallow is a model obtained by performing continual pre-training on Llama with a large Japanese corpus to enhance Japanese language capabilities.

\paragraph{Prompts}
We conduct experiments using nine different prompts.\footnote{The experiments were conducted in May and June 2025.}
We create the prompts based on the templates in the FLAN collection~\citep{Longpre2023TheFC}, which compiles instruction tuning data and methods.
The templates contain multiple evaluation instructions, so we use them to examine the models' robustness to prompts.
The details of the prompts are shown in \Cref{apd:prompts}.

\paragraph{Logical Inference System}
We also evaluate \systemname{}~\cite{ccgjcomp}, a logical inference system for Japanese comparatives.
This system derives semantic representations of the input sentences and performs theorem proving to judge the entailment relation.

\subsection{Results and Discussion}
\label{sec:results1}

\begin{figure}[t]
    \centering
    \includegraphics[width=\linewidth]{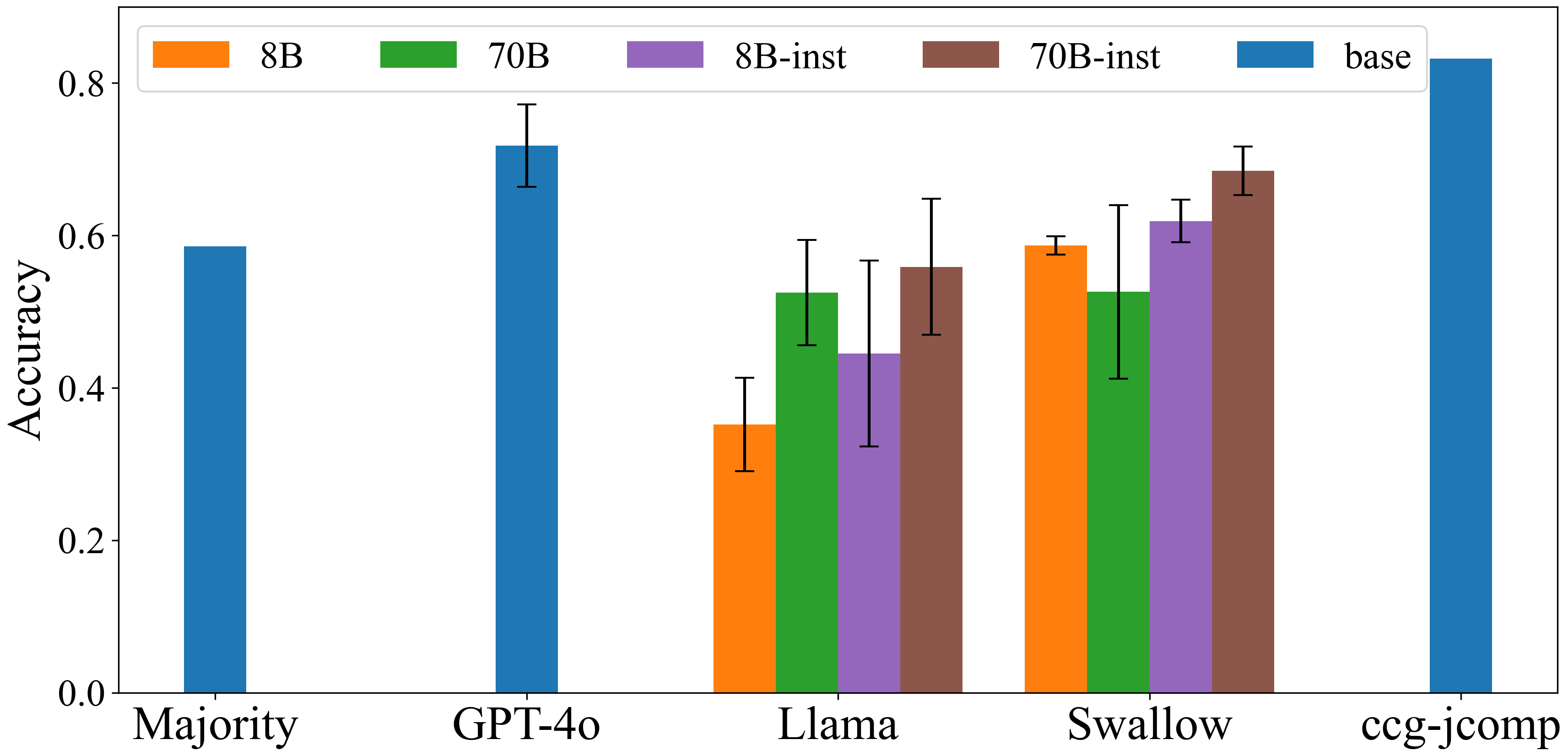}
    \caption{Accuracies on our dataset in the zero-shot setting (average and standard deviation of nine prompts). ``Majority'' indicates the accuracy achieved by answering \textit{yes}, the most frequent label in the dataset, for all problems.}
    \label{fig:results}
\end{figure}

\Cref{fig:results} presents the accuracy of each system.
As shown, GPT-4o demonstrated the best performance of all the LLMs.
Among the open-source models, Swallow, which specifically targets Japanese, outperformed Llama.
In addition, larger models performed better, and instruction-tuned models outperformed their non-tuned counterparts of the same size.
All models had variations depending on the prompt, and these variations were particularly large for Llama8B-inst and Swallow70B.

LLMs tended to produce incorrect answers even for relatively simple problems.
For instance, they often incorrectly answered \textit{yes} to the problems generated from jsem-570 in \Cref{tab:dataset_example}, possibly due to the lexical overlap between the premise and the hypothesis.
Previous studies have suggested that there are lexical overlap heuristics or order-preserving subset heuristics in pre-trained models performing NLI tasks~\citep{mccoy-etal-2019-right, yanaka-mineshima-2021-assessing}.
The experimental result indicates that such heuristics may also be present in LLMs.

We also highlight that the LLMs struggled to handle linguistic phenomena that exist in Japanese but not in English.
GPT-4o failed to correctly answer the problems related to presupposition (e.g., jsem-620), which is unique to Japanese comparatives.
About Llama and Swallow, they tended to incorrectly answer \textit{yes} to problems such as (\ref{ex:or-comparative}), in which (\ref{ex:or-comparative-1}) is the premise and (\ref{ex:or-comparative-2}) is the hypothesis.
\begin{exe}
    \ex\label{ex:or-comparative}
    \begin{xlist}
        \ex\label{ex:or-comparative-1}
        \gll Taro-wa Jiro ka Saburo-yori omoi. \\
                 Taro-\Top{} Jiro or Saburo-than heavy\\
        \glt ``Taro is heavier than Jiro or Saburo.''
        \ex\label{ex:or-comparative-2}
        \gll Taro-wa Jiro-yori omoi. \\
             Taro-\Top{} Jiro-than heavy\\
        \glt ``Taro is heavier than Jiro.''
    \end{xlist}
\end{exe}

\noindent
Here, the gold label is \textit{unknown} because the disjunction in (\ref{ex:or-comparative-1}) cannot have narrow scope below \textit{than}.
In contrast, its English counterpart does allow such a reading (i.e., Taro is heavier than both Jiro and Saburo), making the label \textit{yes}. 
It is possible that the errors of the models are due to this difference between the two languages.

\section{Evaluation of Few-shot NLI}
\label{chap:fewshot}
Next, we analyze the extent to which model predictions change depending on how few-shot examples related to the problem category are given.

\subsection{Experimental Setting}
\label{sec:settings2}
For GPT-4o, Llama70B-inst, and Swallow70B-inst, we conduct two types of few-shot experiments with the prompt that showed the highest accuracy in \Cref{chap:zeroshot}.

\paragraph{Few\_normal}
For each problem, we give the models one few-shot example generated from the same template.
For instance, we show an example generated from jsem-570 to a model, and then evaluate it on a modified version where at least one of X, Y, and A is replaced with a different word.

\paragraph{Few\_adversarial}
For each problem, we give the models an example that is closely related to the problem but has a different gold label.
For example, when evaluating a model on jsem-577, we give it an example whose premise is augmented with ``Y-wa A'' (Y is A).
This revision changes the gold label to \textit{yes}.
Note that we conduct this experiment only for categories with more than one kind of gold label.

\newcommand{\normal}{\textsc{few}\_\textsc{normal}}
\newcommand{\adversarial}{\textsc{few}\_\textsc{adversarial}}

\subsection{Results and Discussion}
\label{sec:results2}

\Cref{fig:results_few_shot} shows the accuracies of the three models in each setting.
In \normal, all the models showed improved accuracy compared to the zero-shot setting.
In particular, Swallow70B-inst exhibited a significantly larger improvement than the other two.
In \adversarial, the accuracy of GPT-4o showed a slight improvement, whereas Llama70B-inst and Swallow70B-inst exhibited performance degradation, which was especially notable in Swallow70B-inst.

The results of the two experiments indicate that Swallow70B-inst is highly susceptible to the gold labels of few-shot examples.
The other two models effectively leveraged the few-shot examples with the same label, and also were not greatly affected when given examples with a different label.

Although the models avoided many of the errors in the zero-shot experiment with the prompts in \normal, the accuracy did not improve sufficiently in some cases.
For example, GPT-4o still failed to correctly answer the problems that require an understanding of presuppositions.
In addition, the accuracy of Llama70B-inst for the problems such as (\ref{ex:or-comparative}) was zero.

\newcommand{\orange}[1]{\textcolor{orange}{#1}}
\begin{figure}
    \centering
    \includegraphics[width=\linewidth]{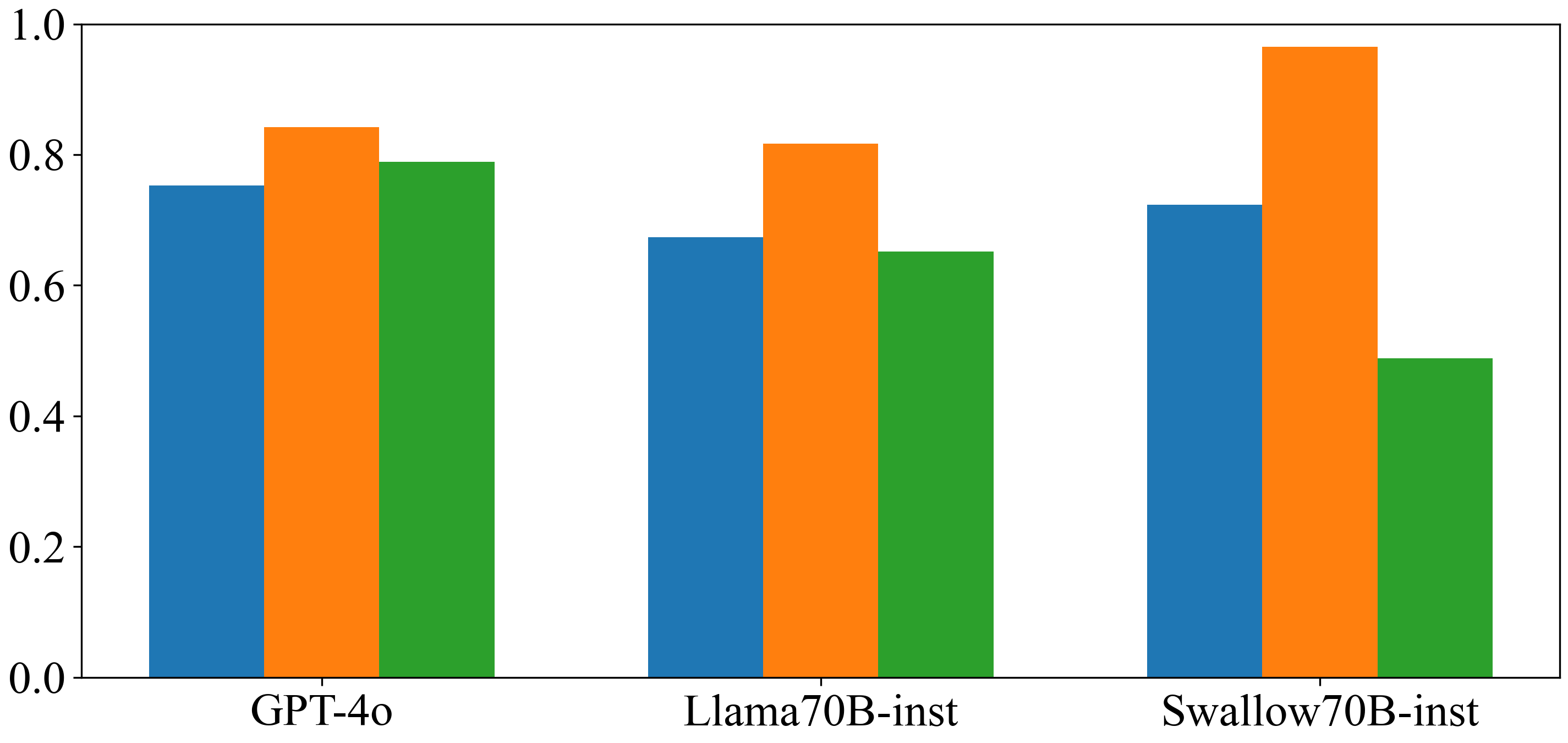}
    \caption{Accuracies of three LLMs in each experimental setting (\blue{blue: zero-shot}; \orange{orange: \normal{}}; \green{green: \adversarial{}})}
    \label{fig:results_few_shot}
\end{figure}

\subsection{Analysis with Semantic Representation Prompts}
\label{subsec:semantic_representation_prompts}
Inspired by \citet{ozeki-etal-2024-exploring}, we construct few-shot prompts with not only example problems, but also their semantic representations obtained via \systemname~(see \Cref{apd:prompt_SR} for details).
We instruct LLMs to generate semantic representations of sentences and then infer the entailment label.
We conduct experiments on problems with which each model showed low accuracy even with the \normal{} prompt: namely, presupposition (e.g., jsem-620) for GPT-4o and disjunctive sentences (e.g., (\ref{ex:or-comparative})) for Llama70B-inst.
As a result, the accuracy of GPT-4o and Llama70B-inst increased from 0.049 to 0.230 and from 0.0 to 0.148, respectively.
This result suggests that providing semantic representations can improve model performance.

\section{Conclusion}
\label{chap:conclusion-and-future-work}

In this study, we constructed an NLI dataset focusing on Japanese comparatives, and analyzed how robustly LLMs can perform inference involving comparatives in zero-shot and few-shot settings.
The zero-shot experiment revealed that the models' performance varies depending on the prompts, and each model exhibited a distinctive pattern of errors.
In the few-shot experiments, we observed that some models, such as Swallow70B-inst, showed a decrease in accuracy when given adversarially designed examples.
This observation suggests that some models may be overly sensitive to the specific labels included in the few-shot examples.
For problems that the models struggled to solve in the few-shot settings, we found that the accuracy can be improved by making the models predict the semantic representations of the sentences.

\section*{Acknowledgments}
We thank the three anonymous reviewers for their helpful comments and feedback. This work was partially supported by the Institute for AI and Beyond of the University of Tokyo, and JSPS KAKENHI grant number JP24H00809.

\bibliographystyle{acl_natbib}
\bibliography{latex/myref}

\clearpage
\appendix

\section{Prompt Templates}
\label{apd:prompts}
\Cref{tab:prompts} shows the prompt templates used in \Cref{chap:zeroshot,chap:fewshot}.
They are translations of the templates in FLAN related to NLI.
\begin{table*}
    \centering
    \small
    \begin{tabular}{p{7.5cm}p{7.5cm}}
    \hline
    \begin{tabular}{p{7cm}}
        Template
    \end{tabular}
    &
    \begin{tabular}{p{7cm}}
        Translation
    \end{tabular}
    \\
    \hline\hline
        \begin{tabular}{p{7cm}}
            \{premises\}\\
            選択肢付きの質問です：上記の段落に基づいて「\{hypothesis\}」と結論付けることはできますか。\\
            選択肢：含意、矛盾、中立\\
            回答：
        \end{tabular}
        &
        \begin{tabular}{p{7cm}}
            (\{premises\}\\
            Question with options: Based on the paragraph above can we conclude that ``\{hypothesis\}''?\\
            options: entailment, contradiction, neutral\\
            answer:)
        \end{tabular}
        \\
    \hline
        \begin{tabular}{p{7cm}}
            \{premises\}\\
            この段落に基づいて、下の文が真であると結論付けることはできますか。\\
            \{hypothesis\}\\
            選択肢：含意、矛盾、中立\\
            回答：
        \end{tabular}
        &
        \begin{tabular}{p{7cm}}
            (\{premises\}\\
            Based on that paragraph can we conclude that the sentence below is true?\\
            \{hypothesis\}\\
            options: entailment, contradiction, neutral\\
            answer:)
        \end{tabular}
        \\
    \hline
        \begin{tabular}{p{7cm}}
            \{premises\}\\
            選択肢付きの質問です：以下の結論を導くことはできますか。\\
            \{hypothesis\}\\
            選択肢：含意、矛盾、中立\\
            回答：
        \end{tabular}
        &
        \begin{tabular}{p{7cm}}
            (\{premises\}\\
            Q with options: Can we draw the following conclusion?\\
            \{hypothesis\}\\
            options: entailment, contradiction, neutral\\
            answer:)
        \end{tabular}
        \\
    \hline
        \begin{tabular}{p{7cm}}
            \{premises\}\\
            前の文が与えられたとき、この次の文は従いますか。\\
            \{hypothesis\}\\
            選択肢：含意、矛盾、中立\\
            回答：
        \end{tabular}
        &
        \begin{tabular}{p{7cm}}
            (\{premises\}\\
            Does this next sentence follow, given the preceding text?\\
            \{hypothesis\}\\
            options: entailment, contradiction, neutral\\
            answer:)
        \end{tabular}
        \\
    \hline
        \begin{tabular}{p{7cm}}
            \{premises\}\\
            選択肢：含意、矛盾、中立\\
            問題：次の文を推論できますか。\\
            \{hypothesis\}\\
            回答：
        \end{tabular}
        &
        \begin{tabular}{p{7cm}}
            (\{premises\}\\
            options: entailment, contradiction, neutral\\
            Question: Can we infer the following?\\
            \{hypothesis\}\\
            answer:)
        \end{tabular}
        \\
    \hline
        \begin{tabular}{p{7cm}}
            次の段落を読んで仮説が真かどうかを決定してください。最後の選択肢の中から選んでください：\\
            \{premises\}\\
            仮説：{hypothesis}\\
            選択肢：含意、矛盾、中立\\
            回答は
        \end{tabular}
        &
        \begin{tabular}{p{7cm}}
            (Read the following paragraph and determine if the hypothesis is true. Select from options at the end:\\
            \{premise\}\\
            Hypothesis: \{hypothesis\}\\
            options: entailment, contradiction, neutral\\
            answer:)
        \end{tabular}
        \\
    \hline 
        \begin{tabular}{p{7cm}}
            テキストを読んで文が真かどうかを決定してください：\\
            \{premises\}\\
            文：\{hypothesis\}\\
            選択肢：含意、矛盾、中立\\
            回答：
        \end{tabular}
        &
        \begin{tabular}{p{7cm}}
            (Read the text and determine if the sentence is true:\\
            \{premises\}\\
            Sentence: \{hypothesis\}\\
            options: entailment, contradiction, neutral\\
            answer:)
        \end{tabular}
        \\
    \hline 
        \begin{tabular}{p{7cm}}
            選択肢付きの質問です：以下の文脈から仮説を導くことはできますか。\\
            文脈：\\
            \{premises\}\\
            仮説：\{hypothesis\}\\
            選択肢：含意、矛盾、中立\\
            回答：
        \end{tabular}
        &
        \begin{tabular}{p{7cm}}
            (Question with options: can we draw the following hypothesis from the context?\\
            Context:\\
            \{premises\}\\
            Hypothesis: \{hypothesis\}\\
            options: entailment, contradiction, neutral\\
            answer:)
        \end{tabular}
        \\
    \hline
        \begin{tabular}{p{7cm}}
            次の文が真かどうかをその下のテキストに基づいて決定してください。選択肢から選んでください。\\
            \{hypothesis\}\\
            \{premises\}\\
            選択肢：含意、矛盾、中立\\
            回答：
        \end{tabular}
        &
        \begin{tabular}{p{7cm}}
            (Determine if the sentence is true based on the text below. Choose from options.\\
            \{hypothesis\}\\
            \{premises\}\\
            options: entailment, contradiction, neutral\\
            answer:)
        \end{tabular}
        \\
    \hline
    \end{tabular}
    \caption{Prompt templates used in \Cref{chap:zeroshot}}
    \label{tab:prompts}
\end{table*}

\section{Results by Category in Zero-shot Experiments}
\begin{figure*}
    \centering
    \includegraphics[width=\linewidth]{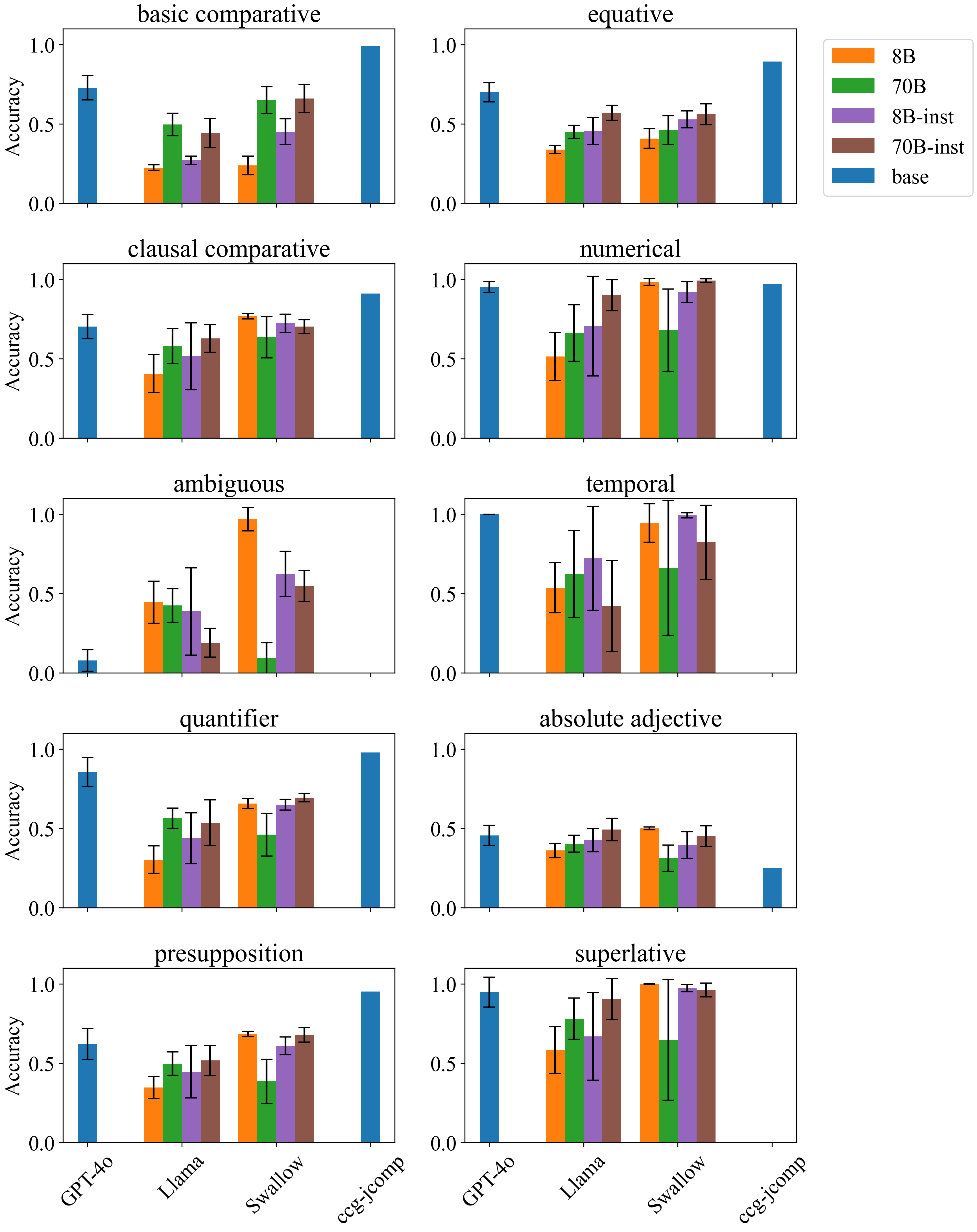}
    \caption{Accuracies of each model and system across categories.}
    \label{fig:category_results}
\end{figure*}

\Cref{fig:category_results} shows the accuracies of each LLM and \systemname~across categories.

\section{Errors of LLMs in the Zero-shot Experiments}
\label{apd:error_analysis}
In addition to the errors described in \Cref{sec:results1}, the LLMs also failed to correctly answer the problems related to equatives such as jsem-577-1 in \Cref{tab:dataset_example}.
They tended to answer \textit{no}, which suggests that they interpret the premise as meaning that the degrees of the two people are exactly equal.

\section{Details of the Experiment with Semantic Representation Prompts}
\label{apd:prompt_SR}
\begin{table*}
    \centering
    \small
    \begin{tabular}{l}
    \hline
        与えられた前提と仮説の間の正しい論理関係を決定してください。\\
        - 仮説が前提から論理的に導かれる場合は「含意」と答えてください。\\
        - 前提と仮説が論理的に両立しない場合は「矛盾」と答えてください。\\
        - 「含意」でも「矛盾」でもない場合は「中立」と答えてください。\\
        \\
        \#\# 入力\\
        前提：太郎は次郎か三郎より明るい。\\
        仮説：太郎は次郎より明るい。\\
        \\
        \#\# 述語論理への翻訳\\
        前提：∃d (明るい(太郎, d) ∧ ￢明るい(次郎, d)) ∨ ∃d (明るい(太郎, d) ∧ ￢明るい(三郎, d))\\
        仮説：∃d (明るい(太郎, d) ∧ ￢明るい(次郎, d))\\
        \\
        \#\# 推論 \\
        \lbrackその答えに対する理由を説明してください\rbrack\\
        \\
        \#\# 回答\\
        \lbrack含意、矛盾、中立のいずれかを答えてください\rbrack\\
        \\
        (Determine the correct logical relationship between the given premises and the hypothesis.\\
        - Answer ``entailment'' if the hypothesis follows logically from the premises.\\
        - Answer ``contradiction'' if the premises and the hypothesis are logically incompatible with each other.\\
        - Answer ``neutral'' if the relationship is neither ``entailment'' nor ``contradiction.''\\
        \\
        \#\# Input\\
        Premise: Taro is more cheerful than Jiro or Saburo.\\
        Hypothesis: Taro is more cheerful than Jiro.\\
        \\
        \#\# Translation into predicate logic\\
        Premise: ∃d (cheerful(taro, d) ∧ ￢cheerful(jiro, d)) ∨ ∃d (cheerful(taro, d) ∧ ￢cheerful(saburo, d))\\
        Hypothesis: ∃d (cheerful(taro, d) ∧ ￢cheerful(jiro, d))\\
        \\
        \#\# Reasoning\\
        \lbrack Explain your reasoning for the answer \rbrack\\
        \\
        \#\# Answer\\
        \lbrack Answer entailment, contradiction, or neutral \rbrack)\\
    \hline
    \end{tabular}
    \caption{
        Instruction and a few-shot example used in the experiment in \Cref{subsec:semantic_representation_prompts}.
        The lower part is an English translation.}
    \label{tab:prompt_SR}
\end{table*}

\Cref{tab:prompt_SR} shows the instruction and a few-shot example used in \Cref{subsec:semantic_representation_prompts}.
It provides the semantic representations adopted in \systemname.

\begin{table*}
    \centering
    \begin{tabular}{l|l}
        \hline
        Input &
        \begin{tabular}{l}
            前提：太郎は次郎か三郎より優しい。\\
            仮説：太郎は次郎より優しい。\\
            (Premise: Taro is kinder than Jiro or Saburo.\\
            Hypothesis: Taro is kinder than Jiro.)
        \end{tabular}
        \\
        \hline
        \begin{tabular}[c]{@{}l@{}}
            Semantic\\
            Representations
        \end{tabular}
        &
        \begin{tabular}{p{11cm}}
            前提：∃d(優しい(太郎, d) ∧ ￢優しい(次郎, d)) ∨ ∃d(優しい(太郎, d) ∧ ￢優しい(三郎, d))\\
            仮説：∃d(優しい(太郎, d) ∧ ￢優しい(次郎, d))\\
            (Premise: ∃d(kind(taro, d) ∧ ￢kind(jiro, d)) ∨ ∃d(kind(taro, d) ∧ ￢kind(saburo, d))\\
            Hypothesis: ∃d(kind(taro, d) ∧ ￢kind(jiro, d)))
        \end{tabular}
        \\
        \hline
        Reasoning
        &
        \begin{tabular}{p{11cm}}
           前提が真であれば、太郎は次郎より優しい、または太郎は三郎より優しい。仮説が真であれば、太郎は次郎より優しい。したがって、仮説は前提から論理的に導かれる。 \\
           (If the premise is true, then Taro is kinder than Jiro, or Taro is kinder than Saburo. If the hypothesis is true, then Taro is kinder than Jiro. Therefore, the hypothesis is logically derived from the premise.)
        \end{tabular}
        \\
        \hline
    \end{tabular}
    \caption{Example of reasoning errors of Llama70B-inst. Semantic representations and reasoning are the output.}
    \label{tab:reasoning_incorrect}
\end{table*}

As for the experimental results, although the accuracy of Llama 70B Instruct was still low compared to other models, the semantic representations it predicted were correct in most problems.
Most of the errors stemmed from the reasoning step.
\Cref{tab:reasoning_incorrect} is an example of reasoning errors.
The semantic representations are correct; the model successfully interpreted the premise as ``Taro is kinder than Jiro, or Taro is kinder than Saburo.''
However, it incorrectly concluded that the hypothesis follows the premise.

\end{document}